\documentclass[runningheads]{llncs}
\usepackage{graphicx}
\usepackage{amsmath}
\usepackage{caption}
\usepackage{subcaption}
\usepackage{multirow}
\usepackage{xcolor}
\usepackage{sidecap}
\usepackage{hyperref}

\makeatletter
\newcommand{\printfnsymbol}[1]{%
  \textsuperscript{\@fnsymbol{#1}}%
}
\makeatother

\begin{document}
\title{TabAug: Data Driven Augmentation for Enhanced Table Structure Recognition}
\titlerunning{TabAug: Data Driven Augmentation}
%
%
\author{Umar Khan\thanks{These authors have equal contribution.} \and
Sohaib Zahid\printfnsymbol{1} \and
Muhammad Asad Ali \and\newline
Dr. Adnan ul-Hasan \and
Dr. Faisal Shafait}
\authorrunning{U. Khan et al.}

\institute{National University of Sciences and Technology (NUST) , H-12 Islamabad, Pakistan 
}
\maketitle              
%

\begin{abstract}
Table Structure Recognition is an essential part of end-to-end tabular data extraction in document images. The recent success of deep learning model architectures in computer vision remains to be non-reflective in table structure recognition, largely because extensive datasets for this domain are still unavailable while annotating new data is expensive and time-consuming. Traditionally, in computer vision, these challenges are addressed by standard augmentation techniques that are based on image transformations like color jittering and random cropping. As demonstrated by our experiments, these techniques are not effective
for the task of table structure recognition. In this paper, we propose TabAug, a re-imagined Data Augmentation technique that produces structural changes in table images through replication and deletion of rows and columns. It also consists of a data-driven probabilistic model that allows control over the augmentation process. To demonstrate the efficacy of our approach, we perform experimentation on ICDAR 2013 dataset where our approach shows consistent improvements in all aspects of the evaluation metrics, with cell-level correct detections improving from $92.16\%$ to $96.11\%$ over the baseline.
\end{abstract}

\keywords{ table structure recognition \and table augmentation \and data augmentation \and table data extraction \and probabilistic model \and data-driven model \and table segmentation \and deep learning.}

\section{Introduction}
\label{sec:introduction}
\vspace{-0.1cm}

Document structure analysis and parsing are some of the most crucial parts of document image processing for digitization and information extraction. One of the most important components of documents is tables. Tables are a structured way of representing data allowing for visual and logical grouping of data in a highly comprehensible manner. Tables in documents are frequently used to present key information such as financial records, receipts, and data forms. Extracting this information can be vital and beneficial to most businesses around the world. However, tabular data extraction can be a challenging task as more often this data is found as scanned document images that contain no structural or content metadata.

Tabular data extraction consists of three major tasks, Table Detection, Table Structure Recognition, and Semantic Understanding.
Table detection is the task of locating tables in a given document, while table structure recognition aims towards the segmentation of tables into rows and columns for layout understanding. Semantic understanding involves the assignment of information such as row or column header (e.g., Unit price, Stock value, etc.) to a particular cell. Our focus in this work is the table structure recognition part, where the aim is to extract rows and columns in a given table image.


In practice, Convolutional Neural Network (CNN) is used as an effective tool for extracting meaningful features from visual information. Given that scanned document images contain only visual information, CNNs become increasingly important for Table Detection and Table Structure Recognition alike. State-of-the-art CNNs are efficient at extracting visual features from data; however, they can be highly data demanding in nature. This precondition of CNNs coupled with the unavailability of large Table Structure Recognition datasets presents a challenging problem as annotating large datasets can be both expensive and time-consuming. It is, therefore, crucial to focus on methods for improving the data-efficiency of deep learning models to achieve better results for Table Structure Recognition on smaller datasets.

Augmentation techniques are widely used to improve the data efficiency in Deep Neural networks. It is the practice of adding slight realistic changes to the original data to increase the diversity of the training data. This concept helps model regularize and avoid over-fitting in small datasets. There are several image transformation techniques such as scaling, rotating, smearing, etc that are widely used in Computer Vision to augment data. For the purposes of comparison, we choose to apply Random cropping, (first employed in AlexNet~\cite{nips_imagenetcnn_12}) and Color Jitter (used in re-implementation of ResNet~\cite{reimp_resnet_16} by Facebook AI Research) as the \emph{standard} augmentation. Please refer to the bottom row of Figure~\ref{fig:aug_samples} for example of these transformations on ICDAR 2013 dataset. However, as shown by our experimentation in Section~\ref{sec:experiments and results}, they are not effective for table structure recognition as they do not alter the structure of a table but instead produce unrepresentative data, which results in a decreased performance.

\begin{figure}[tbp]
    \centering
    \parbox{\textwidth}{
        \parbox{.2\textwidth}{%
            \subcaptionbox{Column Replication\label{fig:col_rep}}
                [.8\linewidth]{
                        \includegraphics[width=.2\textwidth]{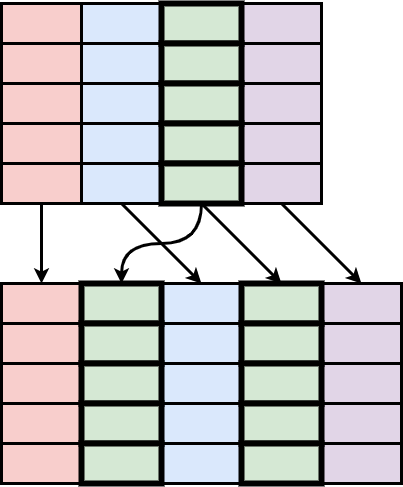}
                }
        }
        \hfill
        \parbox{.2\textwidth}{%
            \subcaptionbox{Column Deletion\label{fig:col_del}}
                [.8\linewidth]{
                        \includegraphics[width=.2\textwidth]{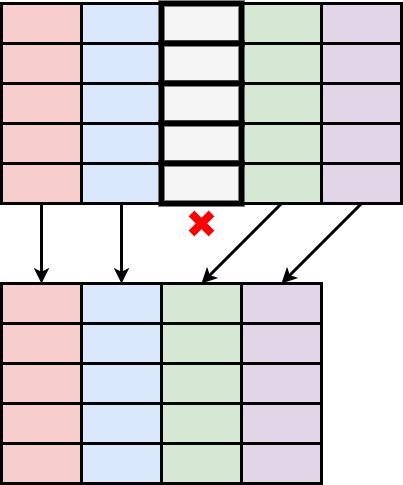}
                }
        }
        \hfill
        \parbox{.4\textwidth}{%
            \subcaptionbox{Row Replication\label{fig:row_rep}}
                [.8\linewidth]{
                        \includegraphics[width=.4\textwidth]{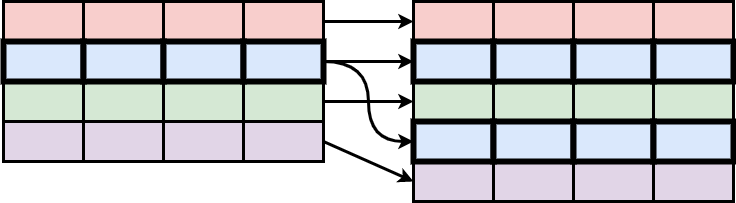}
            }
            \vskip1em
            \subcaptionbox{Row Deletion\label{fig:row_del}}
                [.8\linewidth]{
                        \includegraphics[width=.4\textwidth]{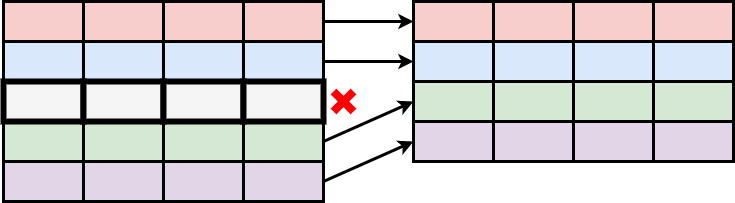}
            }
        }
    }
        \caption{Visual depiction of proposed augmentation operations}
        \label{fig:ops}
\end{figure}

To overcome these challenges, we propose a novel method of data augmentation based on two key operations of \emph{Replication} and \emph{Deletion} on rows and columns, illustrated in Figure~\ref{fig:ops} and sample images of these illustrated operations can be seen in Figure~\ref{fig:aug_samples}. We also introduce a data-driven probabilistic model for generating parameters that control the outlook of the augmented data. The augmented variations generated by this technique better represent the real-world variation in tables.
Please refer to Section~\ref{sec:methodolgoy} for further details about our methodology.
Our experiments on ICDAR 2013 dataset demonstrate that our proposed augmentation technique yields a higher data efficiency and out-performs both the non-augmented and standard image augmentation approaches. 
Please refer to Section~\ref{sec:experiments and results} for further details on experimental evaluation.


These results set a strong foundation for more future work in this direction. To facilitate this, we have decided to make our implementation of structural data augmentation named \emph{TabAug} and associated experimentation code open-source\footnote{\textcolor{blue}{\url{https://github.com/sohaib023/splerge-tab-aug}}}. We have also developed the existing code to be fairly simple to plug into various existing models with support for T-Truth annotation format~\cite{shahab_das10} with no training overhead.

The rest of paper is structured into following sections. 
Section~\ref{sec:related_work}, outlines literature review of relevant works in the domain of table structure recognition and tabular data extraction in general. In Section~\ref{sec:methodolgoy} we present our methodology and implementation of the proposed augmentation technique. In Section~\ref{sec:experiments and results} we report on our experiments and comparative results. Finally, in Section~\ref{sec:conclusion} we conclude our research with future direction for our work.


\section{Related Work}
\label{sec:related_work}
\vspace{-0.1cm}
Tabular data extraction is an old and recognized problem with solutions maturing from over past 20 years. Starting with one of the earliest works in table detection in text files using heuristics was done by~\cite{Tupaj1996ExtractingTI} in 1996 followed by Pyreddy et al.~\cite{pyreddy_97} in 1997. The following years saw a novel heuristic-based approach of detecting tables in document images from Keinenger et al.~\cite{kieninger_das98},~\cite{kiegner_icapr99} and~\cite{kieninger_icdar01} forming a combined table plotting and recognition system named TRECS. Zanibbi et al.~\cite{zanibbi_04} presented a comprehensive survey paper that focused on table recognition literature writing down extensive observations on various aspects of popular methods available at the time. A novel approach of detecting tables from ruling lines was presented by Basilios et al.~\cite{basilios_05}. In 2010 Shahab et al.~\cite{shahab_das10} presented comprehensive and rigorous evaluation metrics for table detection and structure recognition. Significant work was done by Shafait et al.~\cite{shafait_das10} in developing table detection algorithm for multiple layouts, they also introduced more meaningful performance metrics for table detection. 

Use of data-driven approaches in table detection started from Chen et al.~\cite{chenjin_11} in 2011 where SVMs were used for table detection in handwritten documents with noise and artifacts. In 2013 Kasar et al.~\cite{kasar_icdar13} also made use of SVMs for table region detection by classifying ruling lines. The following year Anukriti et al.~\cite{anukriti_14} used Conditional Random Fields (CRFs) with encoded foreground block characteristics and the contextual information as features for learning layouts and labeling table and non-table regions in documents.

From 2015 onward, Deep Learning models have been used extensively for solving the detection and recognition problem. ~\cite{gilani_icdar17},~\cite{schreiber_icdar17},~\cite{saman_dicta18} and~\cite{Siddiqui2018DeCNTDD} made use of Faster R-CNN~\cite{FasterRCNN15} with their own take on handcrafted features and pre-processing methodologies. These methods were successful in producing state-of-the-art results for table detection but failed to produce significant results in Structure Recognition. Schreiber et al.~\cite{schreiber_icdar17}  mentions the unnatural approach of detecting rows and columns through Faster R-CNN~\cite{FasterRCNN15} and instead proposes fine-grained image segmentation through FCN-X's architecture by Shelhamer et al.~\cite{shelhamer_ieee_17}, however, their proposed technique heavily relies upon image stretching for expanding background pixels as a pre-processing methodology. 

In more recent works from Qasim et al.~\cite{sharukh_gnn_19} and Tensmeyer et al.~\cite{splitting_merging_19} on table structure recognition, we see a more stable and natural approach towards the formulation of problem. Qasim et al.~\cite{sharukh_gnn_19} made use of Graph Neural Networks for generating cell adjacency matrix for all existing OCR detected words and Tensmeyer et al.~\cite{splitting_merging_19} formulating the problem of structure recognition as a combination of row, column splits and cell mergings for defining the structure of a table. Qasim et al.~\cite{sharukh_gnn_19} mentions that lack of large datasets has been a major hindrance for Deep Learning methods in table structure recognition and makes use of synthetic data to show the effectiveness of their network. However, synthetic data generated randomly is limited in capturing the visual and general characteristics of the target dataset, due to which the model fails to perform on the target dataset despite the optimal results on the synthetic data. Similarly for the split model from Tensmeyer et al.~\cite{splitting_merging_19}, even though more data-efficient than Qasim et al.~\cite{sharukh_gnn_19} GNN model, makes use of large proprietary dataset to train the network for an effective model to be tested on ICDAR 2013.
Our experiments of training and testing of split model from Tensmeyer et al.~\cite{splitting_merging_19} on ICDAR 2013 dataset reveals sub-optimal results than the network potential due to the lack of diverse and large training dataset.

With newer deep learning algorithms, we are seeing a trend of increase in performance on larger datasets; however, their results translate to sub-optimal results on smaller datasets thus limiting their potential use cases. Traditionally in computer vision, these problems are addressed with image transformation based data augmentation techniques defined as standard data augmentation for increasing data diversity in training data. Albeit successful in natural images, standard augmentation hold an insignificant and unstable impact on table images.
In this paper, we propose a new re-imagined Tabular Data Augmentation inspired from~\cite{Dwibedi2017CutPA} and~\cite{Ghiasi2020SimpleCI} by replicating and removing table structure elements (rows and columns) to form augmented tables all while maintaining their visual artifacts, we have also introduced a data-driven probabilistic model (similar to~\cite{fang_instaboost_19}) for decision parameters of our method of Tabular Data Augmentation.


\section{Methodology}
\label{sec:methodolgoy}
\vspace{-0.1cm}
We adopt an elementary approach of structural data augmentation for table structure recognition. We consider cells to be the building block of a table, that is, a table is defined as a combination of cells. Therefore, it should theoretically be possible to achieve a large number of combinations given a small number of cells. However, combining different cells into a table poses many limitations with regards to compatibility. First, widths and heights of cells can vary greatly, and thus combining them without due deliberation can lead to unnatural tables. Second, cells tend to inherit their formatting and styling from their adjacent cells. In a fully randomized collection of cells, one cell might have left-justified text while the next might have right-justified text, making for a confusing and unnatural table.

To overcome these limitations, we redefine a table to be a combination of rows and columns rather than cells. This helps keep intact the co-relation of cell styling within a row and column. To further simplify the problem, we limit these combinations to randomization of rows and columns within the same table. Thus, we can generate a randomized permutation of the rows and columns within a table to obtain a structurally augmented version of the table. This augmentation technique is then combined with a data-driven probabilistic sampler for maximizing its effectiveness.

\vspace{-0.2cm}
\subsection{Augmentation Operations}
\vspace{-0.1cm}
To achieve randomization of rows and columns, we define four basic operations that we can apply to each table:

\begin{figure}[tbp]
    \centering
    \parbox{\textwidth}{
        \parbox{.33\textwidth}{%
            \subcaptionbox{Original\label{fig:aug_org}}
                [\linewidth]{
                    \includegraphics[width=.32\textwidth]{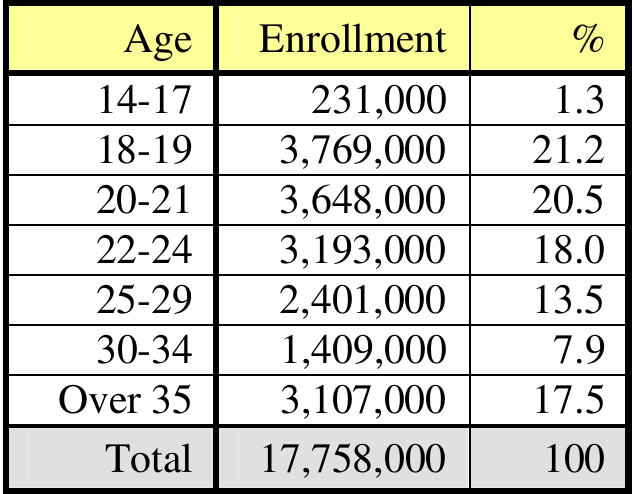}
                }
        }
        \hfill
        \parbox{.66\textwidth}{%
            \subcaptionbox{\label{fig:aug1}}
                [0.45\linewidth]{
                \includegraphics[width=.32\textwidth]{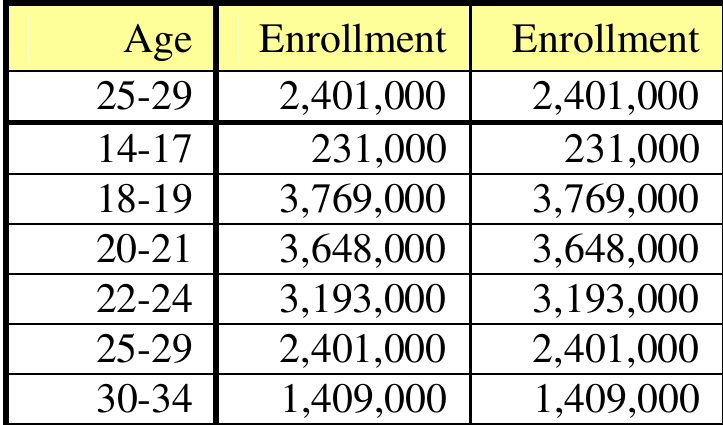}
            }
            \hfill
            \subcaptionbox{\label{fig:aug2}}
                [0.45\linewidth]{
                \includegraphics[width=.32\textwidth]{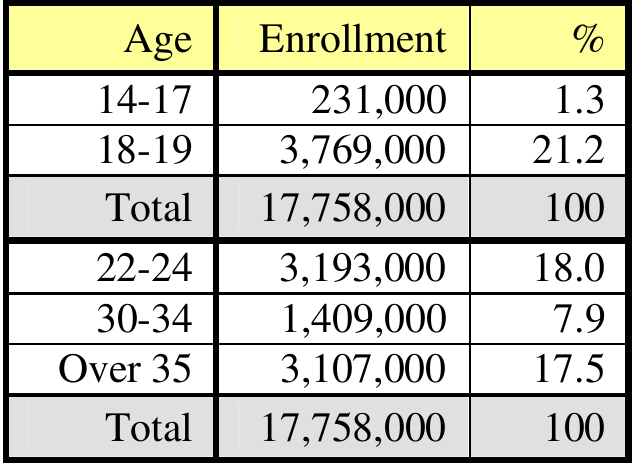}
            }
            \hfill
            
            \subcaptionbox{\label{fig:cls1}}
                [0.45\linewidth]{
                \includegraphics[width=.32\textwidth]{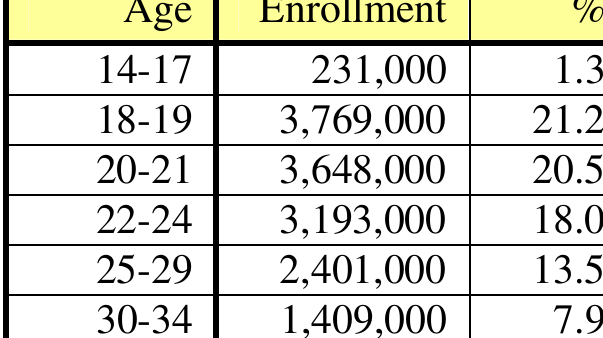}
            }
            \hfill
            \subcaptionbox{\label{fig:cls2}}
                [0.45\linewidth]{
                \includegraphics[width=.32\textwidth]{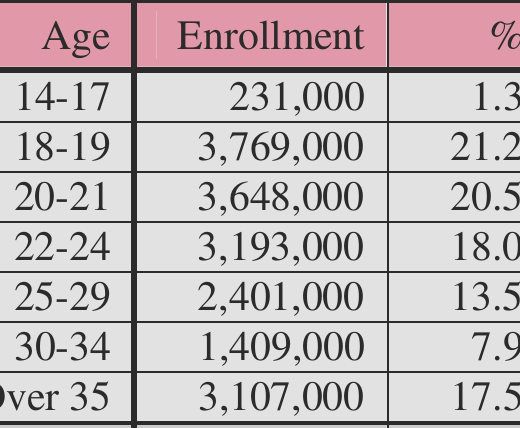}
            }
            \hfill
        }
    }
        \caption{Sub-figure~\ref{fig:aug_org} contains an original table sample from ICDAR 2013 dataset. Sub-figures~\ref{fig:aug1} and~\ref{fig:aug2} show variations of the table generated by TabAug. Sub-figures~\ref{fig:cls1} and~\ref{fig:cls2} show the results of image transformation based augmentations.}
        \label{fig:aug_samples}
\end{figure}

\begin{itemize}
    \item Row deletion
    \item Column deletion
    \item Row replication
    \item Column replication
\end{itemize}
These operations are depicted visually in Figure~\ref{fig:ops} for better understanding. These are the core atomic operations that we utilize during our augmentation pipeline. They can be applied sequentially in random orders on a table to achieve increasingly varying versions of the same table.

\vspace{-0.2cm}
\subsection{Augmentation Sub-Operations}
\vspace{-0.1cm}

Each of the proposed augmentation operations consists of several sub-operations, which are further explained below. For simplicity, we will only consider Column deletion and replication, as Row deletion and replication can be directly inferred from it. 
\vspace{-0.2cm}
\subsubsection{Source Selection:} 
Before doing any replication or removal, we need to select a column on which we may apply the operation. We randomly select an index $c$ in the range of $\{1\,..\, C-1\}$, where $C$ is the total number of columns in a given table. We purposefully leave out the $0-th$ index as the first column can have row headers (similarly column headers in the case of row selection), which should be retained in its location to maintain a natural table. 

Furthermore, in case if spanning cells exist in the selected column, we must ensure that no partial/broken cells are copied.
\begin{equation}
    \text{span}_{\text{min}} = \min(\{\text{cells}_{r\,c}^{\text{Start Column}} \,\, \forall r \in \{0..R-1\}\})
    \label{eq:span_min}
\end{equation}
\begin{equation}
    \text{span}_{\text{max}} = \max(\{\text{cells}_{r\,c}^{\text{End Column}} \,\, \forall r \in \{0..R-1\}\})
    \label{eq:span_max}
\end{equation}
 
\begin{figure}[tbp]

    \subcaptionbox{Non-convex column selection.
        \label{fig:convex-a}
    }
        [.3\linewidth]{
            \includegraphics[width=.25\textwidth]{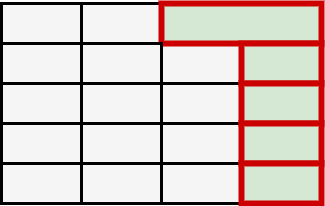}
    }
    \hfill
    \subcaptionbox{Expansion of selection to retrieve convex block.
        \label{fig:convex-b}
    }
        [.3\linewidth]{
            \includegraphics[width=.25\textwidth]{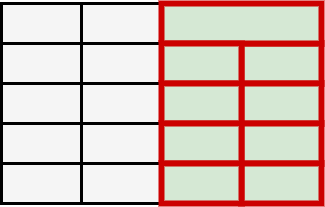}
    }
    \hfill
    \subcaptionbox{Example where expansion fails to retrieve convex block.         \label{fig:convex-c}
    }
        [.3\linewidth]{
            \includegraphics[width=.25\textwidth]{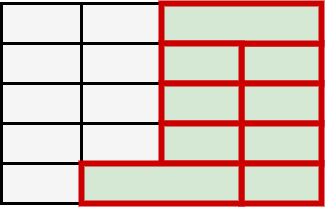}
    }
    \caption{A non-convex selection~\ref{fig:convex-a} cannot be replicated or deleted and hence it is expanded so as to make it convex~\ref{fig:convex-b}. Sub-figure~\ref{fig:convex-c} represents the case where expansion also results in a non-convex selection. 
    }
    \label{fig:convex}
\end{figure}

For the selected column $c$, the values of $\text{span}_{\text{min}}$ and $\text{span}_{\text{max}}$ are calculated using equations~\ref{eq:span_min} and~\ref{eq:span_max}. These values provide us with column indices for a self contained convex block (as depicted in Figure~\ref{fig:convex-a} \& ~\ref{fig:convex-b}). In case if even after performing the above steps the selected block is non-convex  (as depicted in Figure~\ref{fig:convex-c}) we abort the operation and return the table unaltered. Otherwise our selection is defined by $c_{\text{max}}=\text{span}_{\text{max}}$ and $c_{\text{min}}=\text{span}_{\text{min}}$ which is used by next sub-operations.

\vspace{-0.2cm}
\subsubsection{Target Location Selection (only for replication):}
This sub-operation is similar to the previous one with few modifications. We select an index $d$ in the range of $\{1\,..\, C\}$ randomly, where $C$ is the total number of columns in a given table. The target location for column replication will be at the start of $d$-th column. So we purposefully leave out the $0-th$ index as we do not want to move the first column. Furthermore, if $d=C$ then the target location is after the last column at the end of the image. We perform a check to see if the target location $d$ is intersecting with a spanning cell. If so, placing a column at that location will cause the existing spanning cell to split apart, and hence we change $d$ according to the equations below:

\begin{equation}
    d = 
        \begin{cases}
        \text{span}_{\text{min}}, & \text{if abs}(d - \text{span}_{\text{min}}) <= \text{abs}(d - \text{span}_{\text{max}})\\ & \text{and span}_{\text{min}} \neq 0\\
        \text{span}_{\text{max}} + 1, & \text{otherwise}
        \end{cases}
\end{equation}
where $\text{span}_{\text{min}}$ and $\text{span}_{\text{max}}$ are calculated using equations~\ref{eq:span_min} and~\ref{eq:span_max} by setting $c=d$.

Similar to the previous sub-operation, if after the correction of $d$ it intersects with a spanning cell we abort the operation and return the table unaltered.
\vspace{-0.2cm}
\subsubsection{Execution:}
To execute the operation, whether deletion or replication, we need to alter both the table image and the ground-truth of the table, the specifics of which vary between the two operations. Following values are required in both operations:

\[
   x_{\text{min}} = \text{columns}[c_{\text{min}}].x1\,\,\,\,\,\,,\,\,\,\,\,\,
   x_{\text{max}} = \text{columns}[c_{\text{max}}].x2\,\,\,\,\,\,,\,\,\,\,\,\,
   w = x_{\text{max}} - x_{\text{min}}
 \]
 where $x1$ denotes starting x-coordinate of a column and $x2$ denotes the ending x-coordinate of a column.


\paragraph{Deletion:}
Firstly, we remove the columns with indices in the range ($c_{\text{min}}$, $c_{\text{max}}$) from the ground-truth, including all the contained cells and span information. For the columns and cells having indices greater than $c_{max}$, we subtract $w$ from their x-coordinates, to move them to the left. 

For the image, we cut out the image from $x_{\text{min}}$ to $x_{\text{max}}$ and then move left the image pixels to the right side of $x_{\text{max}}$ by $w$.

\paragraph{Replication:}
Firstly, we calculate the x-coordinate of the target location where the replicated column is to be placed. 
\begin{equation}
    x_{\text{dst}} = \text{columns}[d].x1 
\end{equation} 
Then we copy the columns with indices in the range ($c_{\text{min}}$, $c_{\text{max}}$) from the ground-truth, including all the contained cells and span information. For the copied columns and cells we offset their x-coordinates by $x_{\text{dst}}-x_{\text{min}}$, so as to move it to the target location. Further, the columns and cells having indices greater than $d$, we add $w$ to their x-coordinates, to move them to the right, clearing up space for the replicated column. 

For the image, we move the image pixels to the right of ${x_{\text{dst}}}$ by $w$, so as to clear up space for crop of the replicated columns. Then we copy the image pixels from $x_{\text{min}}$ to $x_{\text{max}}$ and move them by $x_{\text{dst}}-x_{\text{min}}$, so as to move it to destination location $x_{\text{dst}}$.

\vspace{-0.2cm}
\subsection{Augmentation Pipeline}
\vspace{-0.1cm}
Having defined the augmentation operations, an effective pipeline must be established that can utilize the proposed operations for training a model, therefore we propose a pipeline in the following.
\begin{figure}[tp]
    \centering
    \includegraphics[width=0.675\textwidth]{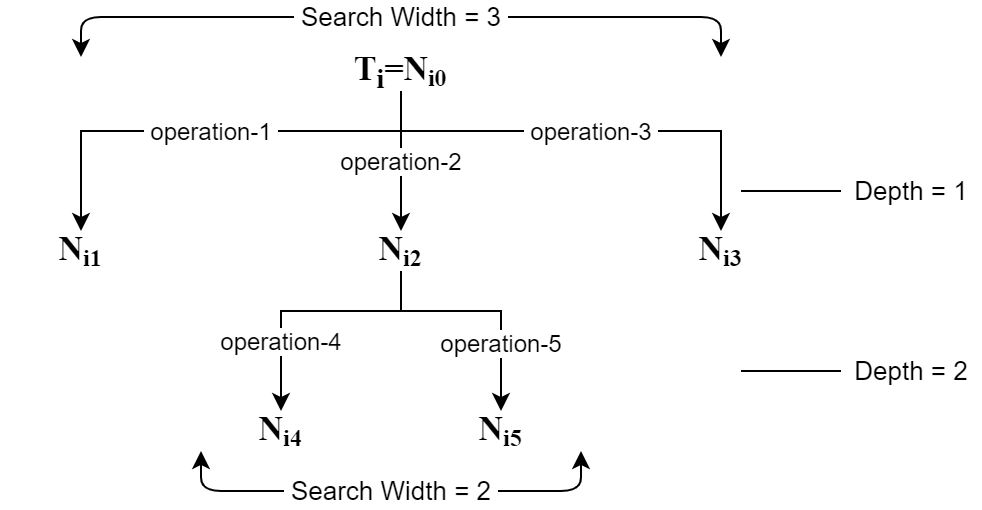}
    \caption{An example of an augmentation search tree. Each of the nodes $N_{ij}$ is obtained by applying an augmentation operation to its parent node, where the root of the tree is the original table $T_i$.}
    \label{fig:tree}
\end{figure}


\vspace{-0.3cm}
\subsubsection{Augmentation Tree Exploration:}
Before training a model, we pre-compute a sample set $N_i$ of augmented versions of a given table $T_i$. Each member of $N_{i}$ is considered to be a node (represented as $N_{ij}$), which is achievable by applying a series of augmentation operations (edges in the tree). An example of such a partial tree is shown in Figure~\ref{fig:tree}. By applying this tree search, we can extract a controlled sample set of all achievable augmented tables of a given root table $T_i$. 
To avoid inundation of samples, we apply pruning to the tree search to keep the number within a reasonable limit. 
First, only the nodes having tree depth greater than 5 and less than 10 are kept as a part of this sample set. Further, the maximum search width for each depth level is also restricted for pruning purposes. For depth=1, maximum search width is 8, for depth=2, it is 4, for depth=\{3..5\} it is 2, and finally for depth=\{6..10\} it is 1. Furthermore, out of the obtained sample set, any node is ignored that has a pixel height or width greater than 1.5 times the original table $T_i$, to ensure unnaturally large tables are not used for training.


\vspace{-0.3cm}
\subsubsection{Categorization of Tables and their Nodes:}
\begin{figure}[tbp]
    \centering
    \includegraphics[width=0.5\textwidth]{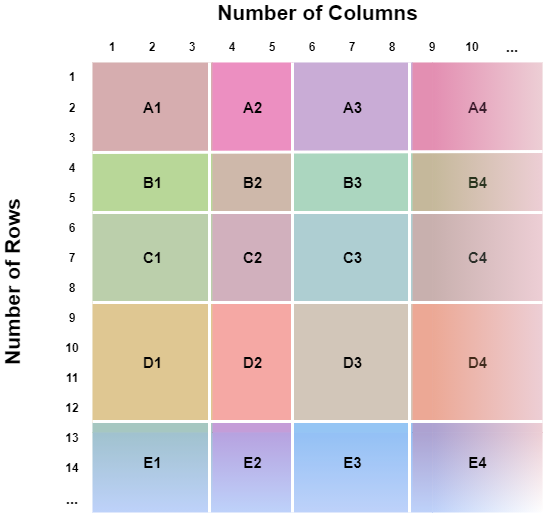}
    \caption{A grid enumerating all of the possible table categories. Each table is assigned one of these categories based on its number of rows and columns}
    \label{fig:table_cat}
\end{figure}

As a pre-processing step, both the tables and their nodes are mapped into categories based on their number of rows and columns. The mapping table is of size 5 x 4 as depicted in Figure~\ref{fig:table_cat}. An example of such a mapping is that a node with 4 rows and 5 columns will be assigned to $B2$ category. Further, if the number of columns in $N_{ij}$ are greater than 10, it is still assigned category $\{A-E\}4$. Similarly, if the number of rows in $N_{ij}$ are greater than 14, it is assigned category $E\{1-4\}$.

After the completion of this step each $T_i$ and $N_{ij}$ will have a category associated with it which is referenced as $T_i^{\text{cat}}$ and $N_{ij}^{\text{cat}}$ respectively.
It must be noted that the decision of category boundaries is purely empirical. 

\vspace{-0.3cm}
\subsubsection{Probability Based Selection:}
For each table $T_i$ we construct a probability distribution $P_i$ using which $T_i$'s nodes will be sampled during training. Firstly, a global frequency distribution of the tables over table categories is generated, giving us $F^G$ a 5 x 4 matrix. Further, for each table $T_i$, a frequency distribution of all $N_{ij}$ over the table categories is generated, giving us $F_i$, which is another 5 x 4 matrix. Lastly, we generate another 5 x 4 matrix $gauss_i$ which is a 2-D Gaussian centered around $T_i^{\text{cat}}$. The spread of this Gaussian distribution allows control over the diversity of the nodes $N_{ij}$ that are sampled during training. These 3 matrices are multiplied to obtain $P_i=\text{Gauss}_i * F^G * F_i$.

 During training, a random table category is selected using $P_i$ as the probability distribution. Once the category is selected, one node is randomly sampled from all the nodes of $T_i$ having the selected category. This selected node is passed for training.

\section{Experiments and Results}
\label{sec:experiments and results}

\vspace{-0.1cm}
To evaluate the efficacy of our proposed augmentation methodology, we train the Split model proposed by~\cite{splitting_merging_19} on ICDAR 2013 dataset. We train the model using three different methodologies:

\begin{enumerate}
    \item {\bfseries Non-Augmented}: Images fed for training without any modification.
    \item {\bfseries Standard}: Basic image transformations, such as, hue saturation and brightness jitter in combination with image cropping.
    \item {\bfseries TabAug:} Our proposed augmentation methodology.
\end{enumerate}

The dataset has a total of $128$ pages and $156$ tables, which is divided into train, test and validation splits with a proportion of $0.72:0.2:0.08$ respectively. 
We train the model with 4 different percentages of the training set, that is, $25\%,$ $50\%$, $75\%$ and $100\%$. It must be noted that this is a ratio of the total training set utilized while training and not a ratio of training samples divided by total samples. 
Further, we repeat each experiment $3$ times to get a better estimate of the average results and their deviation. 

For training, we use a batch size of $1$, and a learning rate of $0.00075$. After every $15$ epoch, we decay the learning rate by a factor of $0.8$. Lastly, we train the models for a total of $27,500$ iterations, which we empirically found to be enough for convergence of the model.




\vspace{-0.2cm}
\subsection{Ground Truth}
\vspace{-0.1cm}
For ICDAR 2013 dataset, ground-truths are provided as bounding boxes of cells, along with their starting and ending row/column indices. This ground-truth format was converted to T-Truth 
format~\cite{shahab_das10}, however, the resulting row and column separators were not aligned with the ruling lines of the table, hence, we re-annotated the dataset using T-Truth to align the separators and then cross-checked it to ensure that the annotations were coherent. The models are trained and tested on this re-annotated dataset, however, as they have been manually cross-checked, the evaluation is applicable to original annotations as well. 


While T-Truth annotation format is used for the augmentation process, Split model~\cite{splitting_merging_19} requires pixel-wise annotations. 
To generate such ground-truth, the separators of T-Truth annotations are expanded to the nearest words (horizontally in the case of column ground-truth, and vertically in the case of row ground-truth).

\vspace{-0.2cm}
\subsection{Performance Measures}
\vspace{-0.1cm}
\label{sec:perf_measures}
We chose the evaluation metrics proposed by Shahab et al.~\cite{shahab_das10} primarily for two reasons. First, these metrics are comprehensive, painting a complete picture of a model's performance. Second, these are general-purpose metrics and can be applied to any type of segment such as rows, columns, and cells.

In the evaluation metrics proposed by Shahab et al.~\cite{shahab_das10}, first the ground truth segments and the predicted segments are numbered. A correspondence matrix of shape $n$ x $m$ is created where $n$ is the number of ground truth segments while $m$ is the number of predicted segments. Each entry $[i, j]$ in the matrix stores the number of pixels that are intersecting between the given ground-truth segment $G_i$ and predicted segment $S_j$, that is $|G_i \cap S_j|$. Further, the sum of $i$-th row in the correspondence matrix gives the total number of pixels in ground-truth segment $G_i$ and the sum of $j$-th column gives the total number of pixels in predicted segment $S_j$. Once this correspondence matrix is generated, we define the following evaluation metrics using the threshold value of $T=0.1$ (consequently $1-T=0.9$):


\begin{enumerate}
\item{
    \textbf{Correct Detections}: The total number of ground-truth segments that have a one-to-one mapping with a predicted segment and a major overlap. Concretely, a given ground-truth segment $G_i$ is considered to be a correct detection if it has a major overlap with a predicted segment $S_j$ and $S_j$ does not have a significant overlap with any other ground-truth segment $(G_k;\forall k \neq i)$. That is:
    \[\frac{|G_i \cap S_j|}{{G_i}} > 1 - T \text{ and } \frac{|G_k \cap S_j|}{{G_k}} < T \ \ ;\forall k \neq i\]
}
\item{
    \textbf{Over-Segmentations}: The total number of ground-truth segments that have a significant overlap with more than one predicted segment. That is, a ground-truth segment $G_i$ is over-segmented if:
    \[T < \frac{|G_i \cap S_j|}{{G_i}} < 1 - T \text{ and } T < \frac{|G_i \cap S_k|}{{G_i}} < 1 - T \ \ \text{where}; k \neq j\]
}
\item{
    \textbf{Under-Segmentations}: Total number of predicted segments that have a significant overlap with more than one ground-truth segments. That is, a predicted segment $S_j$ is under-segmented if:
    \[T < \frac{|G_i \cap S_j|}{{G_i}} < 1 - T \text{ and } T < \frac{|G_k \cap S_j|}{{G_k}} < 1 - T \ \ \text{where}; k \neq i\]
}

\end{enumerate}
\vspace{-0.3cm}
All of these evaluation metrics are normalized by division with the total number of ground-truth segments, which are presented as percentages in the results. We intentionally leave out the Partial Detections, Missed Segments, and False Positive Detections from our evaluation metrics, as due to the problem formulation of Split Model~\cite{splitting_merging_19} they always evaluate to zero.

\begin{figure}
     \centering

         \begin{subfigure}[b]{\textwidth}
             \centering
             \includegraphics[width=\textwidth]{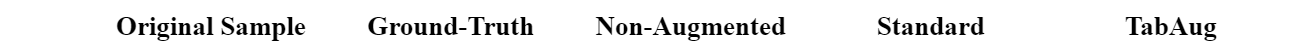}
             \label{fig:analysisheader}
         \end{subfigure}
         \begin{subfigure}[b]{\textwidth}
             \centering
             \includegraphics[width=\textwidth]{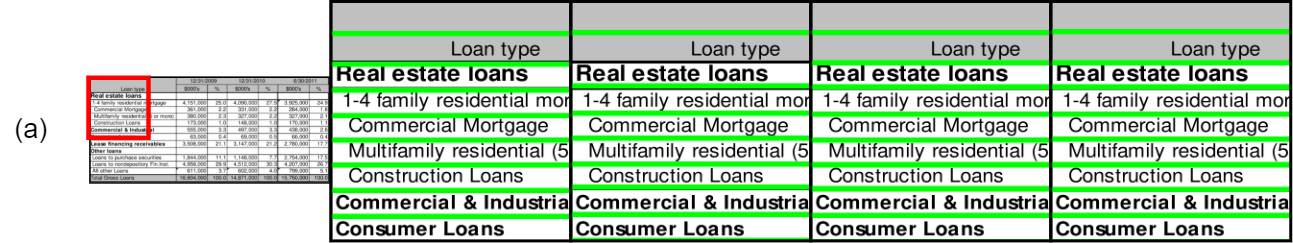}
             \label{fig:analysis1}
         \end{subfigure}
         \begin{subfigure}[b]{\textwidth}
             \centering
             \includegraphics[width=\textwidth]{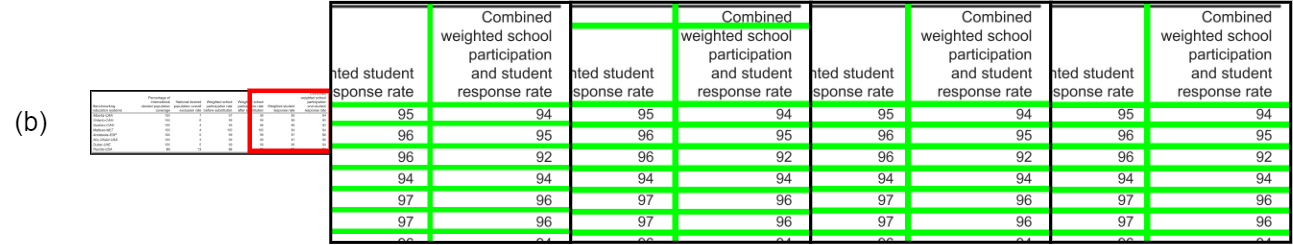}
             \label{fig:analysis2}
         \end{subfigure}
         \begin{subfigure}[b]{\textwidth}
             \centering
             \includegraphics[width=\textwidth]{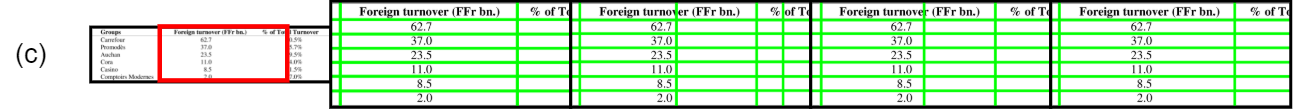}
             \label{fig:analysis3}
         \end{subfigure}
         \begin{subfigure}[b]{\textwidth}
             \centering
             \includegraphics[width=\textwidth]{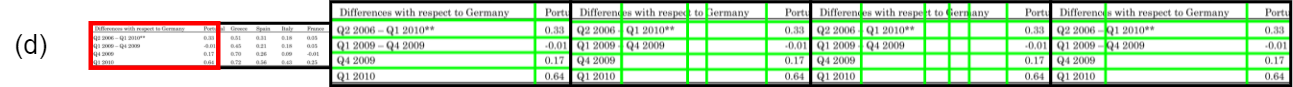}
             \label{fig:analysis4}
         \end{subfigure}
         \begin{subfigure}[b]{\textwidth}
             \centering
             \includegraphics[width=\textwidth]{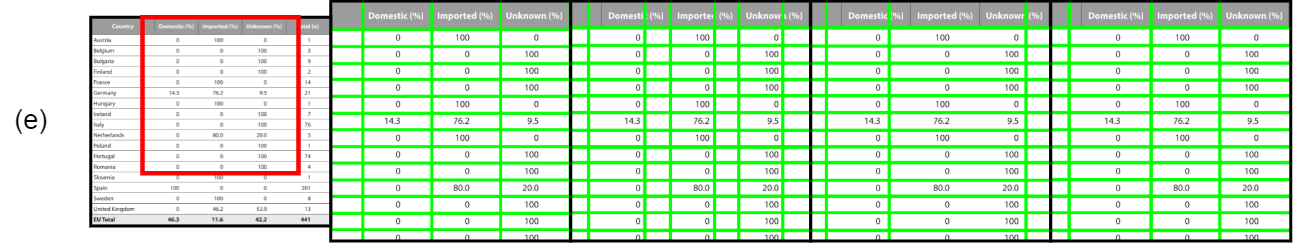}
             \label{fig:analysis5}
         \end{subfigure}
         \begin{subfigure}[b]{\textwidth}
             \centering
             \includegraphics[width=\textwidth]{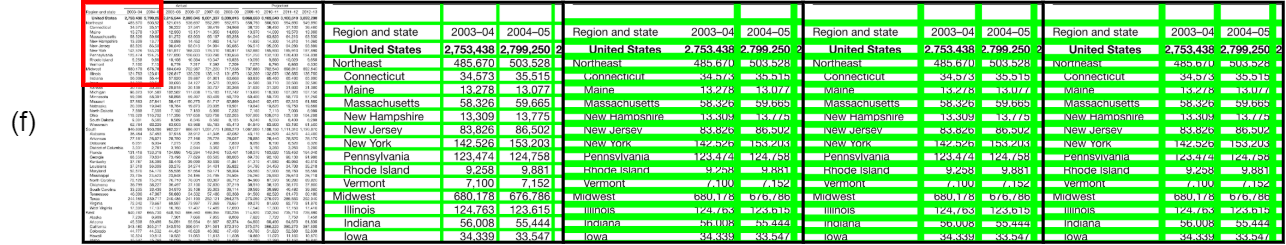}
             \label{fig:analysis6}
         \end{subfigure}
         \begin{subfigure}[b]{\textwidth}
             \centering
             \includegraphics[width=\textwidth]{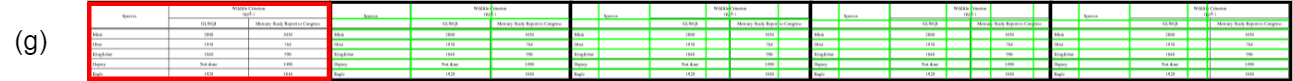}
             \label{fig:analysis7}
         \end{subfigure}
        \caption{Sample outputs of all three approaches for a comparative analysis. Each row provides a separate test sample. The red boxes highlight the image region that is displayed for visualization of ground-truth and the predictions. Ground-truth is displayed as blue lines while predictions are displayed as green lines.}
        \label{fig:analysis}
\end{figure}

\setlength{\tabcolsep}{5pt}
\def\arraystretch{1.1}%

\begin{table}[htbp]
    \caption{Results of evaluating the trained models on 20\% of ICDAR 2013 dataset reserved for testing.}\label{table:eval}
    \begin{center}
    \begin{tabular}{|c|c|c|c|c|}
        \hline
        \,\, && Non-Augmented & Standard & TabAug\\
        \hline
        \hline
        \multirow{3}{4em}{Row} 
        &\,\,Correct  (\%) & 96.44 $\pm$ 1.13  & 97.86 $\pm$ 0.80    & \textbf{97.86 $\pm$ 0.35}\\
        \cline{2-5}
        &\,\,Over-Segmented (\%) & 2.64 $\pm$ 0.54    & 1.71 $\pm$ 0.46    & \textbf{1.71 $\pm$ 0.35}\\
        \cline{2-5}
        &\,\,Under-Segmented (\%) & 0.71 $\pm$ 0.56   & 0.21 $\pm$ 0.18    & \textbf{0.21 $\pm$ 0.18}\\
        \hline
        \hline
        \multirow{3}{4em}{Column} 
        &\,\,Correct    (\%)     & 92.12 $\pm$ 1.11 & 86.38 $\pm$ 1.54  &\textbf{94.44 $\pm$ 0.25}\\
        \cline{2-5}
        &\,\,Over-Segmented (\%)  & 4.84 $\pm$ 0.76  & 5.2 $\pm$ 1.11    & \textbf{3.77 $\pm$ 0.76}\\
        \cline{2-5}
        &\,\,Under-Segmented (\%) & 1.43 $\pm$ 0.25  & 3.95 $\pm$ 1.01   & \textbf{0.9 $\pm$ 0.25}\\
        \hline
        \hline
        \multirow{3}{4em}{Cell} 
        &\,\,Correct   (\%)     & 92.16 $\pm$ 3.84  & 82.12 $\pm$ 6.76  & \textbf{96.11 $\pm$ 1.61}\\
        \cline{2-5}
        &\,\,Over-Segmented (\%)  & 1.58 $\pm$ 0.25  & 1.13 $\pm$ 0.28   & \textbf{1.02 $\pm$ 0.37}\\
        \cline{2-5}
        &\,\,Under-Segmented (\%) & 3.37 $\pm$ 1.99  & 6.9 $\pm$ 2.21    & \textbf{1.46 $\pm$ 0.72}\\
        \hline
    \end{tabular}
    \end{center}
\end{table}

\begin{figure}[t]
    \subcaptionbox{Rows\label{fig:graph_row}
    }
        [.45\linewidth]{
            \includegraphics[width=.4\textwidth]{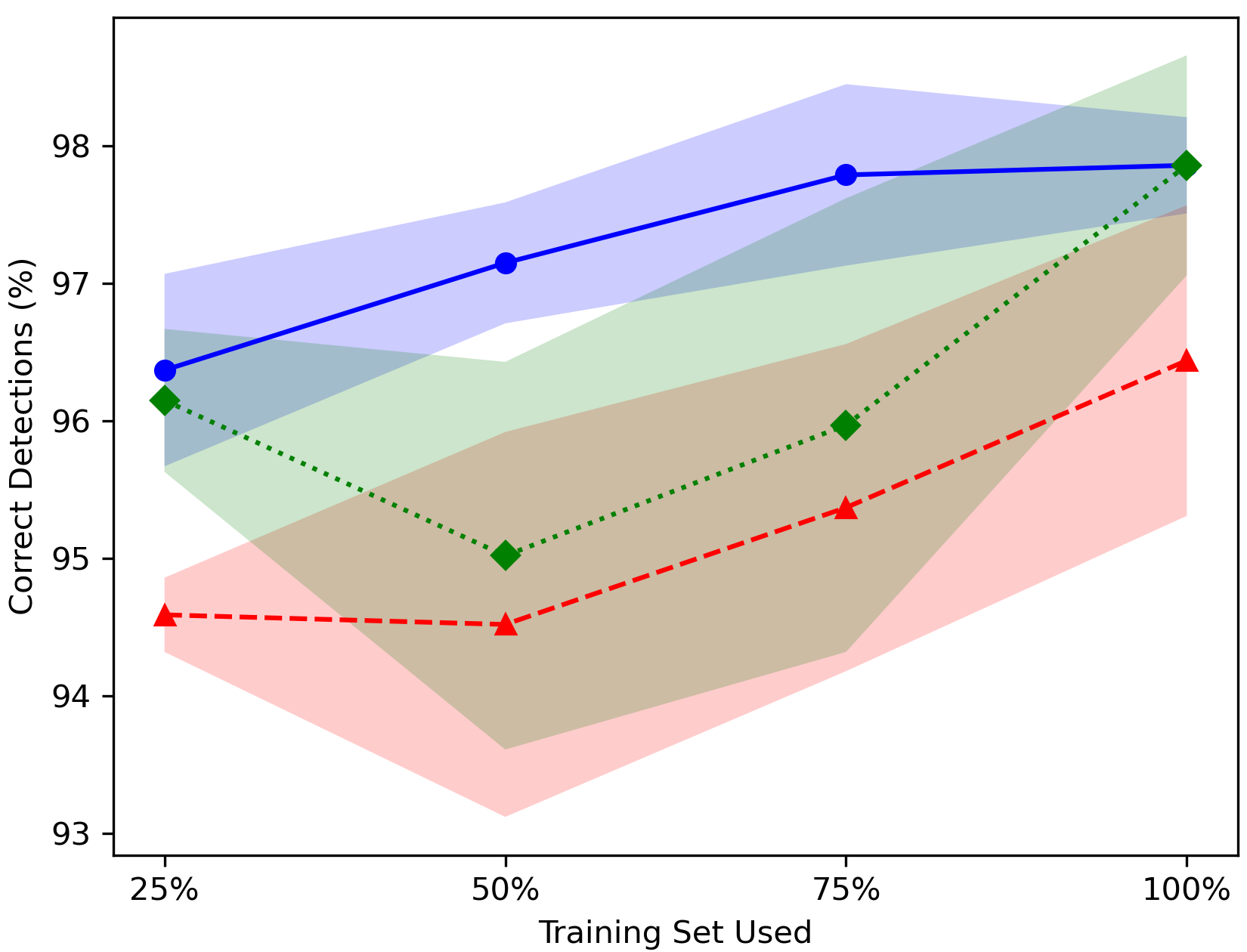}
    }
    \hfill
    \subcaptionbox{Columns\label{fig:graph_col}
    }
        [.45\linewidth]{
            \includegraphics[width=.4\textwidth]{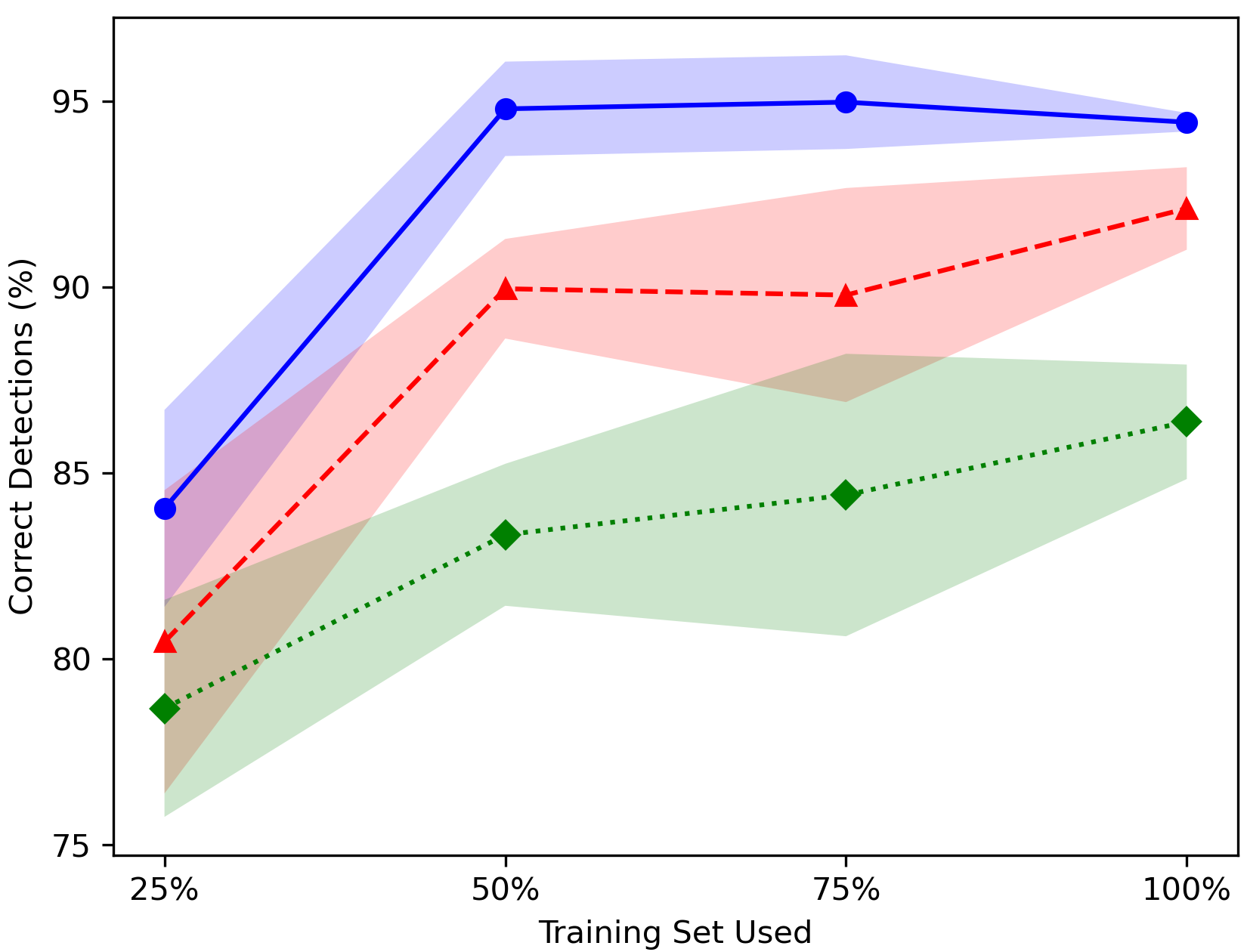}
    }
    \hfill
    \subcaptionbox{Cells\label{fig:graph_cell}
    }
        [.45\linewidth]{
            \includegraphics[width=.4\textwidth]{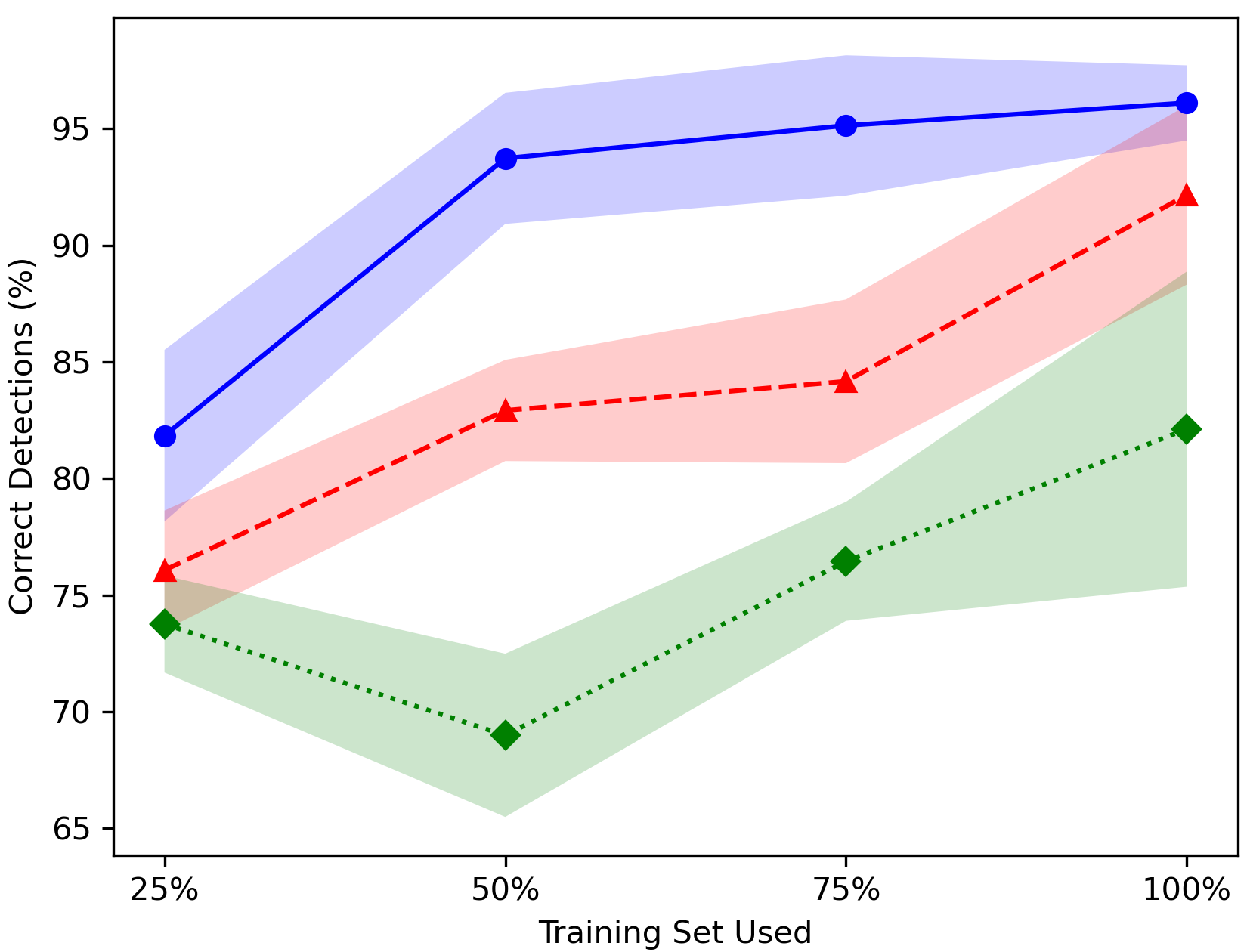}
    }
    \subcaptionbox{Legend\label{fig:graph_legend}
    }
        [.45\linewidth]{
            \includegraphics[width=.3\textwidth]{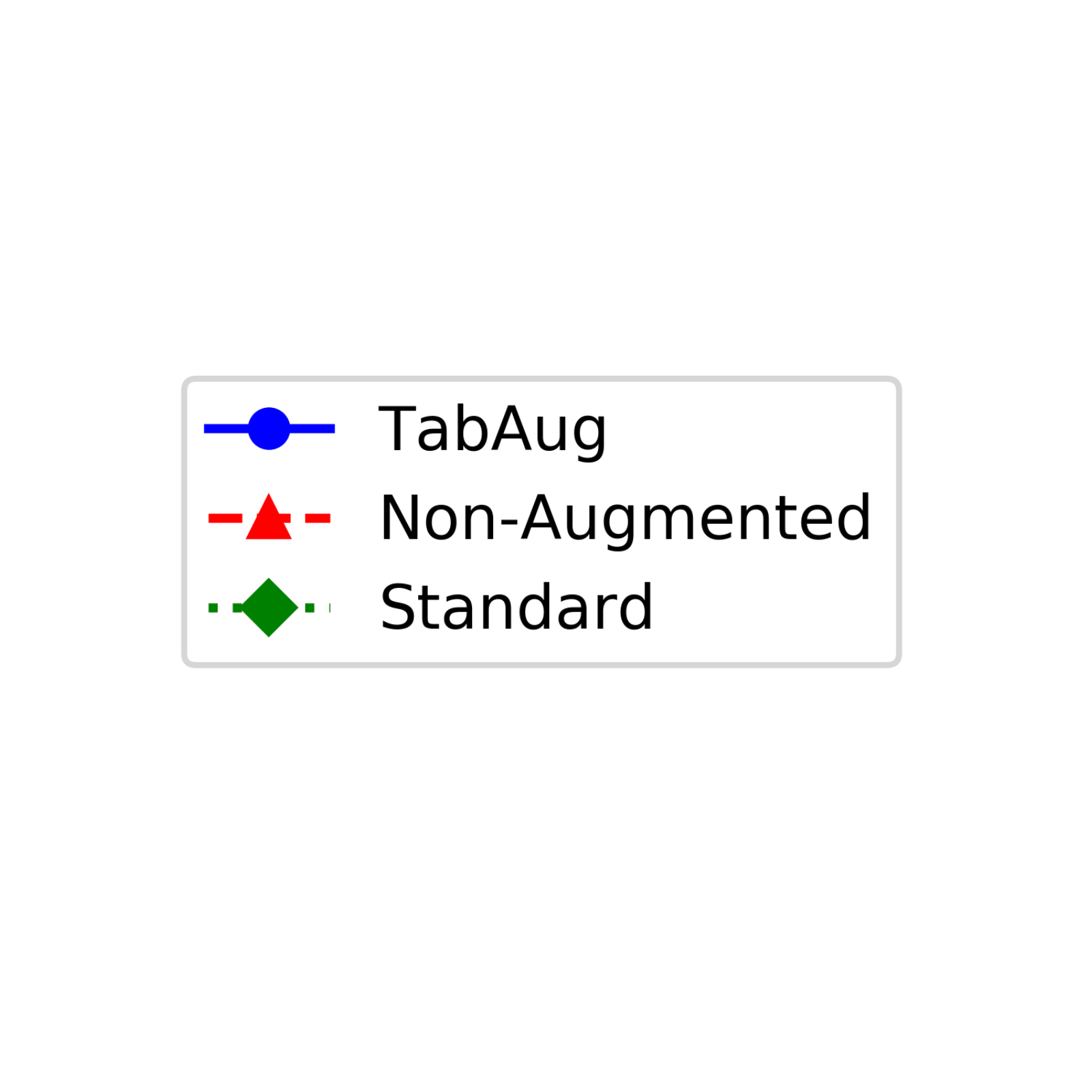}
    }
    \hfill
    \caption{Models were trained on a varying percentage of the total training dataset and then evaluated on a consistent 20\% test split. The graphs show the progression of correct detection with an increase in training data for all three approaches.}
    \label{fig:graph_splits}
\end{figure}

\vspace{-0.2cm}
\subsection{Results and Analysis}
\vspace{-0.1cm}
We evaluate all the trained models on a consistent 20\% test split of ICDAR 2013 dataset. An evaluation of the performance measures described in~\ref{sec:perf_measures} is provided in Table~\ref{table:eval}. Our proposed approach outperforms all the other approaches in all performance measures for each of the Cell, Row, and Column recognition. Further, the models trained with the proposed approach show less standard deviation in the evaluation metrics, which is indicative of more stability in training. Moreover, as depicted in Table~\ref{table:eval_ratios} and Figure~\ref{fig:graph_splits} our proposed approach consistently shows improvements across a range of training dataset sizes.

The standard augmentation approach manages to improve upon the evaluation metrics for Row, however, happens to be deficient in Cell and Column evaluations. This is explained by the fact that Row segmentation, especially in ICDAR 2013 dataset, is an easier task with less visual variability and complexity. Hence, it can be learned mostly based on white-spaces, for which image transformations provide an ample amount of augmentation. Column segmentation contains many complex scenarios and requires learning of more abstract and structural features, for which image transformations turn out to be insufficient.

We have further shown a comparison of the predictions generated by the models trained on each of the three approaches in Figure~\ref{fig:analysis}. In Figure~\ref{fig:analysis}a, we see the case where both standard augmentation and TabAug approaches successfully segment the rows, however, the non-augmented approach under-segments a row in the header region. Conversely, in Figure~\ref{fig:analysis}b, we see a case of over-segmentation by the non-augmented approach. In Figure~\ref{fig:analysis}c, non-augmented and standard augmentation approaches fail to recognize the boundaries of the columns correctly, due to the white-spaces that exist between the protruding words of the header. However, TabAug robustly recognizes the boundaries of the column and predicts a single column separator. In Figure~\ref{fig:analysis}d, we see a scenario where the TabAug approach fails, as it confuses vertically consistent breaks in the text as a column breakage. Regardless, we see that it still does a decent job at making logically sound column segmentations as compared to segmentations from the other two approaches. Figure~\ref{fig:analysis}e depicts another sample that is correctly segmented by TabAug, however causes significant confusion in column segmentation for the other two approaches. Finally, in Figure~\ref{fig:analysis}g the first column is over-segmented by all of the approaches, as the column header has no overlap with its content below. This demonstrates a hard sample that requires a higher cognitive understanding for correct prediction.

\def\arraystretch{1.1}%

\begin{table}[t]
    \caption{Correct detection percentages achieved by models trained on different percentages of the training dataset evaluated on a consistent 20\% test split of ICDAR 2013 dataset.}\label{table:eval_ratios}
    \begin{center}
    \begin{tabular}{|c|c|c|c|c|}
    	\hline
        &&\multicolumn{3}{c|}{Correct Detections (\%)}\\
        \cline{3-5}
        \,\, & Training Data Used & Non-Augmented & Standard & TabAug\\
        \hline
        \hline
        \multirow{4}{4em}{Row} 
        & 25\%      & 94.59 $\pm$ 0.27  & 96.15 $\pm$ 0.52  & \textbf{96.37 $\pm$ 0.7}\\
        \cline{2-5}
        & 50\%      & 94.52 $\pm$ 1.4   & 95.02 $\pm$ 1.41  & \textbf{97.15 $\pm$ 0.44}\\
        \cline{2-5}
        & 75\%      & 95.37 $\pm$ 1.19  & 94.73 $\pm$ 3.38  & \textbf{97.79 $\pm$ 0.66}\\
        \cline{2-5}
        & 100\%     & 96.44 $\pm$ 1.13  & 97.86 $\pm$ 0.8   & \textbf{97.86 $\pm$ 0.35}\\
        \hline
        \hline
        \multirow{4}{4em}{Column} 
        & 25\%      & 80.46 $\pm$ 4.08 & 78.67 $\pm$ 2.92   & \textbf{84.05 $\pm$ 2.65}\\
        \cline{2-5}
        & 50\%      & 89.96 $\pm$ 1.34 & 83.34 $\pm$ 1.91   & \textbf{94.8 $\pm$ 1.27}\\
        \cline{2-5}
        & 75\%      & 89.79 $\pm$ 2.88 & 84.41 $\pm$ 3.8    & \textbf{94.98 $\pm$ 1.26}\\
        \cline{2-5}
        & 100\%     & 92.12 $\pm$ 1.11 & 86.38 $\pm$ 1.54   & \textbf{94.44 $\pm$ 0.25}\\
        \hline
        \hline
        \multirow{4}{3em}{Cell} 
        & 25\%      & 76.07 $\pm$ 2.56 & 73.76 $\pm$ 2.08   & \textbf{81.84 $\pm$ 3.68}\\
        \cline{2-5}
        & 50\%      & 82.92 $\pm$ 2.17 & 68.99 $\pm$ 3.5    & \textbf{93.73 $\pm$ 2.81}\\
        \cline{2-5}
        & 75\%      & 84.17 $\pm$ 3.51 & 76.45 $\pm$ 2.55   & \textbf{95.14 $\pm$ 3.01}\\
        \cline{2-5}
        & 100\%     & 92.16 $\pm$ 3.84 & 82.12 $\pm$ 6.76   & \textbf{96.11 $\pm$ 1.61}\\
        \hline
    \end{tabular}
    \end{center}
\end{table}

\section{Conclusion}
\label{sec:conclusion}
\vspace{-0.1cm}

In this paper, we presented TabAug, a novel augmentation technique capable of producing structural changes in a table through replication and deletion of rows and columns. A data-driven probabilistic model is used in conjunction with the augmentation technique to control the augmentation process. Following the promising results of Split-model~\cite{splitting_merging_19} trained on the publicly available ICDAR 2013 dataset using TabAug, we believe our work provides a strong foundation for numerous future extensions.
In future, we plan to explore ideas for cross-table augmentation through statistical layout matching.

\bibliographystyle{splncs04}
\bibliography{citations}

\begin{thebibliography}{10}
\providecommand{\url}[1]{\texttt{#1}}
\providecommand{\urlprefix}{URL }
\providecommand{\doi}[1]{https://doi.org/#1}

\bibitem{saman_dicta18}
Arif, S., Shafait, F.: Table detection in document images using foreground and
  background features. In: Digital Image Computing: Techniques and Applications
  2018. pp.~1--8 (2018)

\bibitem{anukriti_14}
Bansal, A., Harit, G., Dutta~Roy, S.: Table extraction from document images
  using fixed point model. In: ICVGIP '14: Proceedings of the 2014 Indian
  Conference on Computer Vision Graphics and Image Processing. pp.~1--8 (2014)

\bibitem{chenjin_11}
Chen, J., Lopresti, D.: Table detection in noisy off-line handwritten
  documents. In: 2011 International Conference on Document Analysis and
  Recognition. pp. 399--403. Beijing, China (2011)

\bibitem{Dwibedi2017CutPA}
Dwibedi, D., Misra, I., Hebert, M.: Cut, paste and learn: Surprisingly easy
  synthesis for instance detection. 2017 IEEE International Conference on
  Computer Vision (ICCV) pp. 1310--1319 (2017)

\bibitem{fang_instaboost_19}
{Fang}, H., {Sun}, J., {Wang}, R., {Gou}, M., {Li}, Y., {Lu}, C.: Instaboost:
  Boosting instance segmentation via probability map guided copy-pasting. In:
  2019 IEEE/CVF International Conference on Computer Vision (ICCV). pp.
  682--691 (2019)

\bibitem{basilios_05}
Gatos, B., Danatsas, D., Pratikakis, I., Perantonis, S.J.: Automatic table
  detection in document images. In: Singh, S., Singh, M., Apte, C., Perner, P.
  (eds.) Pattern Recognition and Data Mining. pp. 609--618. Springer Berlin
  Heidelberg, Berlin, Heidelberg (2005)

\bibitem{Ghiasi2020SimpleCI}
Ghiasi, G., Cui, Y., Srinivas, A., Qian, R., Lin, T.Y., Cubuk, E.D., Le, Q.V.,
  Zoph, B.: Simple copy-paste is a strong data augmentation method for instance
  segmentation. ArXiv  (2020)

\bibitem{gilani_icdar17}
Gilani, A., Qasim, S.R., Malik, I., Shafait, F.: Table detection using deep
  learning. In: 14th International Conference on Document Analysis and
  Recognition. pp. 771--776 (2017)

\bibitem{reimp_resnet_16}
{He}, K., {Zhang}, X., {Ren}, S., {Sun}, J.: Deep residual learning for image
  recognition. In: 2016 IEEE Conference on Computer Vision and Pattern
  Recognition (CVPR). pp. 770--778 (2016)

\bibitem{kasar_icdar13}
Kasar, T., Barlas, P., Adam, S., Chatelain, C., Paquet, T.: Learning to detect
  tables in scanned document images using line information. In: Twelfth
  International Conference on Document Analysis and Recognition. pp. 1185--1189
  (2013)

\bibitem{kieninger_das98}
Kieninger, T., Dengel, A.: A paper-to-html table converting system. In:
  Proceedings of document analysis systems. pp. 356--365 (1998)

\bibitem{kiegner_icapr99}
Kieninger, T., Dengel, A.: Table recognition and labeling using intrinsic
  layout features. In: International Conference on Advances in Pattern
  Recognition. pp. 307--316 (1999)

\bibitem{kieninger_icdar01}
Kieninger, T., Dengel, A.: Applying the t-recs table recognition system to the
  business letter domain. In: International Conference on Document Analysis and
  Recognition. p.~0518 (2001)

\bibitem{nips_imagenetcnn_12}
Krizhevsky, A., Sutskever, I., Hinton, G.E.: Imagenet classification with deep
  convolutional neural networks. In: Pereira, F., Burges, C.J.C., Bottou, L.,
  Weinberger, K.Q. (eds.) Advances in Neural Information Processing Systems.
  vol.~25 (2012)

\bibitem{pyreddy_97}
Pyreddy, P., Croft, W.B.: Tinti: A system for retrieval in text tables title2:.
  Tech. rep., University of Massachusetts, USA (1997)

\bibitem{sharukh_gnn_19}
Qasim, S.R., Mahmood, H., Shafait, F.: Rethinking table recognition using graph
  neural networks. In: 2019 International Conference on Document Analysis and
  Recognition (ICDAR). pp. 142--147 (2019)

\bibitem{FasterRCNN15}
Ren, S., He, K., Girshick, R., Sun, J.: Faster r-cnn: Towards real-time object
  detection with region proposal networks. IEEE Transactions on Pattern
  Analysis and Machine Intelligence  \textbf{39} (06 2015).
  \doi{10.1109/TPAMI.2016.2577031}

\bibitem{schreiber_icdar17}
Schreiber, S., Agne, S., Wolf, I., Dengel, A., Ahmed, S.: Deepdesrt: Deep
  learning for detection and structure recognition of tables in document
  images. In: Fourteenth International Conference on Document Analysis and
  Recognition. vol.~1, pp. 1162--1167 (2017)

\bibitem{shafait_das10}
Shafait, F., Smith, R.: Table detection in heterogeneous documents. In:
  Proceedings of the 9th IAPR International Workshop on Document Analysis
  Systems. pp. 65--72. document analysis systems (2010)

\bibitem{shahab_das10}
Shahab, A., Shafait, F., Kieninger, T., Dengel, A.: An open approach towards
  the benchmarking of table structure recognition systems. In: Document
  Analysis Systems. pp. 113--120 (2010)

\bibitem{shelhamer_ieee_17}
Shelhamer, E., Long, J., Darrell, T.: Fully convolutional networks for semantic
  segmentation. IEEE Transactions on Pattern Analysis and Machine Intelligence
  \textbf{39}(4),  640--651 (2017)

\bibitem{Siddiqui2018DeCNTDD}
Siddiqui, S., Malik, M., Agne, S., Dengel, A., Ahmed, S.: Decnt: Deep
  deformable cnn for table detection. IEEE Access  \textbf{6},  74151--74161
  (2018)

\bibitem{splitting_merging_19}
{Tensmeyer}, C., {Morariu}, V.I., {Price}, B., {Cohen}, S., {Martinez}, T.:
  Deep splitting and merging for table structure decomposition. In: 2019
  International Conference on Document Analysis and Recognition (ICDAR). pp.
  114--121 (2019)

\bibitem{Tupaj1996ExtractingTI}
Tupaj, S., Shi, Z., Chang, D.H.: Extracting tabular information from text
  files. In: EECS Department, Tufts University (1996)

\bibitem{zanibbi_04}
Zanibbi, R., Blostein, D., Cordy, J.: A survey of table recognition. IJDAR
  \textbf{7},  1--16 (2004)

\end{thebibliography}
\end{document}